# Title

Analytical reconstructions of full-scan multiple source-translation computed tomography under large field of views

## Author information

*Names and Affiliations of the authors:*

Zhisheng Wang[1,2], Yue Liu[1,2], Shunli Wang[1,2], Xingyuan Bian[1,2], Zongfeng Li[1,2], Junning Cui[1,2,*]

1. *Center of Ultra-precision Optoelectronic Instrument engineering, Harbin Institute of Technology, Harbin, 150080, China*
2. *Key Lab of Ultra-precision Intelligent Instrumentation, Harbin Institute of Technology, Harbin, 150080, Ministry of Industry and Information Technology Harbin, 150080, China*

*Corresponding author:*
Junning Cui
Current address: Building D, Science and Technology Park, Harbin Institute of Technology, Harbin, China.
Email: cuijunning@126.com

**Abstract:** This paper is to investigate the high-quality analytical reconstructions of multiple source-translation computed tomography (mSTCT) under an extended field of view (FOV). Under the larger FOVs, the previously proposed backprojection filtration (BPF) algorithms for mSTCT, including D-BPF and S-BPF (their differences are different derivate directions along the detector and source, respectively), make some errors and artifacts in the reconstructed images due to a backprojection weighting factor and the half-scan mode, which deviates from the intention of mSTCT imaging. In this paper, to achieve reconstruction with as little error as possible under the extremely extended FOV, we combine the full-scan mSTCT (F-mSTCT) geometry with the previous BPF algorithms to study the performance and derive a suitable redundancy-weighted function for F-mSTCT. The experimental results indicate FS-BPF can get high-quality, stable images under the extremely extended FOV of imaging a large object, though it requires more projections than FD-BPF. Finally, for different practical requirements in extending FOV imaging, we give suggestions on algorithm selection.

## 1. Introduction

Micro-computed tomography (Micro-CT) is widely used in the study of object morphology and structure [1–3]. In a lot of scenarios, micro-CT is necessary to image large-sized objects with high resolutions, including larger objects that cannot be cropped, such as some biological samples [4], precious fossils [5], etc., and the demand for high-resolution characterization on a large scale, such as lithium batteries [6,7]. However, the application of micro-CT is restricted due to its limited field of view (FOV) at high spatial resolution [8]. The main reason is that, in the rotating object-based micro-CT, to avoid truncation artifacts, the object should be completely within the FOV, which is the inner circle of the fan-beam formed by the detector and x-ray source. Therefore, the FOV is mainly limited by the size of the detector, whereas the fabrication of large, inexpensive flat panel detectors remains a challenge. Over the decades, several techniques have been developed to enlarge the FOV, including detector offset [9–11], traverse-continuous-rotate scanning [12], rotation-translation-translation multi-scan mode [13], rotation-translation multi-scan mode [14], elliptical trajectory [15], complementary circular scanning [16], and rotated detector [17]. Previously, Yu et al. [18] proposed a source-translation CT (STCT) to enlarge the FOV in an easier way. To acquire the complete projection data set within the FOV, multiple source-translation CT (mSTCT) was further developed with minimum number of STCT segments with different angles [19]. In mSTCT, the measured object is placed close to the micro-focus x-ray source and far from the detector to achieve a large geometric magnification, and the FOV can be adjusted by just changing the length of the source focus translation. This scanning method of extending the FOV inevitably results in each view being transversely truncated, but the projection data set within the FOV is complete [19]. In FBP-type algorithms, each view is filtered individually, and the truncated projection can introduce infinite high-frequency components in the Fourier domain. Because the actual sampling cannot be infinite, the truncated projection after filtering will generate the Gibbs phenomenon at the truncation point.

To solve this problem, iterative algorithms are effective in obtaining high-quality images, but the reconstruction efficiency is not satisfactory in practice [18]. To avoid the truncation, focusing on effective analytical algorithms, Yu et al. [19] designed a virtual projection-based FBP (V-FBP) algorithm for mSTCT, and the virtual projection is the set of measured x-ray divergents for each detector element. V-FBP performs the ramp-filtering for projection along the source, and the cutoff frequency of the filtering is determined by the number of source sampling points. Hence, a large number of projections need to be acquired to ensure the essential spatial resolution. To directly avoid the truncation, another analytical algorithm, i.e., backprojection filtration (BPF), can be introduced into reconstruction for mSTCT, which is mainly divided into differentiated backprojection (DBP) and finite inversion of Hilbert transform in the image domain. Previously, based on the basic concepts of the BPF algorithm, we derived new BPF algorithms for mSTCT [20]. In this study, we utilized the fact that the detector elements are tightly arranged (i.e., small elements with high resolution) to study the loss of less high-frequency information in mSTCT reconstruction with fewer sampling points. More specifically, the derivative is conducted along the detector and source in DBP, obtaining the D-DBP and S-DBP (the letters "D" and "S" indicate the differentiated operation in DBP, which follows the directions of the detector and the source, respectively) formulas, and finally getting two BPF algorithms for mSTCT, i.e., D-BPF and S-BPF, respectively. The derivative is actually equivalent to special ramp filtering, and the resolution of the detector unit is usually higher than the interval between the source sampling points, so the derivative along the detector has a higher cutoff frequency and keep more high-frequency information. Therefore, D-BPF can reconstruct high-quality image with fewer projections, while S-BPF with derivative along the source requires more projections (this disadvantage is the same as V-FBP).

However, we find that in the larger FOVs, D-BPF reconstructions show unstable, intolerable errors in the area near the source, i.e., some streak artifacts distributed at the edges of the image and FOV. The errors become severe as the FOV and object size increase, which goes against the original purpose of developing mSTCT. Additionally, S-BPF does not generate unstable errors at image edges, but artifacts appear in the reconstructed image as the FOV extends, confirming the issue of the previously proposed redundancy weights for mSTCT.

Zeng et al. [21] proposed the theory of reconstruction error in short scanning of circular trajectories. Inspired by this work, we infer that the backprojection weighting factor in the D-DBP formula is a keyshot that causes unstable errors in mSTCT (it is equivalent to a special short-scan mode). In its backprojection weighting factor and the mSTCT configuration with a large geometric magnification, the distance between the reconstructed point and the source may be a very small value. Especially in extended FOV imaging, if the reconstruction point close to the source is not measured twice, this factor will cause very large errors.

Some existing efforts to eliminate the effect of the backprojection weighting factors are mainly focused on standard scanning trajectories (i.e., circular and helical scans), and some strategies mainly include: the circular harmonic reconstruction (CHR) algorithm [22]; rebinning algorithms, i.e., rebinning projection into parallel beams and then reconstruction [23]; the shift-variant filtering methods in the ramp filter based FBP (rFBP) algorithm [24,25]; FBP-type algorithm using the Hilbert transform and one-dimensional coordinate transformation between the parallel-beam and fanbeam coordinates, also called hFBP [26]; introducing additional weighting in full-scan geometry to eliminate weighting factors with a power of one on rays with complementary angles [23,27–29]. Among them, hFBP revealed better stability than the rFBP and CHR algorithms [26]. Under most extended FOV and little redundancy situations, FBP-type algorithms are not better than BPF. The fifth method is proven effective in full-scan standard strategies, but it cannot eliminate the weighting factor with quadratic power in the D-DBP formula. In fact, this backprojection weighting factor with a square term is difficult to eliminate and achieve high-quality reconstruction in half-scan mode [30].

One relatively simple and feasible motivation is to introduce a full-scan mSTCT geometry, namely F-mSTCT, which may weaken the influence of this unstable weighting factor. To achieve this, we first design a redundancy-weighted function that can effectively adapt to F-mSTCT scanning parameters and then introduce it into the previously proposed S-BPF and D-BPF algorithms, obtaining FD-BPF and FS-BPF (the letter "F" refers to F-mSTCT). The reason why we attempt to perform FD-BPF and FS-BPF in large FOVs separately is that both algorithms have their own advantages. FD-BPF can reconstruct high-quality images with a few projections, but the unstable errors are difficult to balance well as the denominator in the backprojection weighting factor may approach zero with the large FOVs. The denominator in the backprojection weighting factor of S-BPF is the distance value from the reconstructed point to the detector, which will be much larger than that of D-BPF under large geometric magnification, which may result in more stable reconstruction results.

In this paper, our work makes contributions as follows: 1) We propose an optimized redundancy weighting function for F-mSTCT in such a way that the direction of the redundancy weighting for projection can be adjusted according to practical needs; 2) Through numerical experiments, we give some suggestions for F-mSTCT algorithm selection for different extended FOV imaging requirements; 3) We conclude that FS-BPF enables F-mSTCT to extremely enlarge its FOV and reconstruct artifact-free images, surpassing the theoretical maximum FOV extension of commonly used circular trajectory detector-bias CT.

This paper is organized as follows. In Section 2, imaging geometries and BPF algorithms for mSTCT are introduced. Section 3 illustrates our proposed method. Section 4 shows experimental results. Finally, the discussion and conclusion are given in Sections 5 and 6.

## 2. Preliminaries

### 2.1 Imaging geometries for mSTCT and F-mSTCT

The mSTCT imaging model is described in Figure 1(a), which is a half-scan mode and consists of multiple STCTs with gradually increasing angles ($\theta_n = \Delta\theta \cdot (n-1)$, $n = 1, 2, \ldots, T$, and $T$ is the minimum number of STCTs) [18]. Similar to the half and full scan standard trajectory, we can also introduce the full-scan mSTCT (F-mSTCT) model, as shown in Figure 1(c).

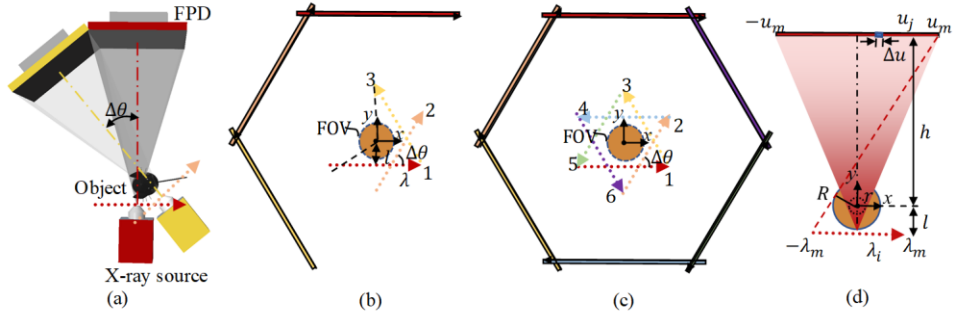

Fig. 1. Imaging models: (a) 3D imaging model mSTCT; (b) and (c) are planer imaging geometries of mSTCT and F-mSTCT, respectively; and (d) illustration of each STCT of (b) and (c). The origin of the fixed coordinate system *o-xyz* is located at the rotated center of the object.

In each STCT, the object is placed close to the source to get a high geometric magnification for the requirement of high resolution, and the source equidistantly and parallelly translates along the row direction of the fixed flat panel detector (FPD) based on the property of the large ray-beam to extend the FOV from the radius $r$ to a larger $R$ (Figure 1(d)). In mSTCT or F-mSTCT, the complete projection data set within the FOV can be acquired, but each view projection is truncated [18].

In this article, we investigate two-dimensional imaging geometry, and the formula of the source trajectory can be expressed as

$$\vec{S}_{\theta_n}(\lambda_i) = [\lambda_i, -l] \cdot \begin{bmatrix} \cos\theta_n & \sin\theta_n \\ -\sin\theta_n & \cos\theta_n \end{bmatrix}, \quad (1)$$

where $\lambda_i$ is the local discrete coordinate of the x-ray source focus ($\lambda$ is its continuous coordinate), $\lambda_i \in [-\lambda_m, \lambda_m]$ ($i$=1, 2, …, $N$, $N$ is the number of source sampling points per STCT), $\lambda_m$ is the half-length of the source translation, and $l$ denotes the origin-to-source distance. $u_j$ is the local discrete coordinate of the x-ray on the detector ($u$ is its continuous coordinate), $u_j \in [-u_m, u_m]$ ($j$=1, 2, …, $J$, $J$ is the number of the detector elements), and $u_m$ is the half-size of the detector. The interval angle $\Delta\theta$ between adjacent STCTs meets the relationship: $2\arctan(u_m/h)$ (see the original paper [18]), and $h$ is the origin-to-detector distance. In F-mSTCT, $T$ can be determined as $T=\lceil 2\pi/\Delta\theta \rceil$, and $\lceil \cdot \rceil$ defines rounding numbers upwards. According to the Ref. [18], the calculation of the radius $R$ of the FOV that can acquire the complete projection data set is

$$R = \frac{\lambda_m h - u_m l}{\sqrt{(l+h)^2 + (\lambda_m + u_m)^2}}. \quad (2)$$

*2.2 BPF-type algorithms for mSTCT*

Previously, our research group proposed BPF algorithms for mSTCT, including D-BPF and S-BPF, to avoid the truncation [20]. Therefore, the D-DBP and S-DBP formulas are derived to obtain different DBP images (see Section Introduction, i.e., intermediate images obtained after DBP operations with different differentiated directions), including $Db_\eta^{\theta_n}(\vec{x})$ and $Sb_\eta^{\theta_n}(\vec{x})$,

$$Db_\eta^{\theta_n}(\vec{x}) = \frac{1}{2} \int_{-\lambda_m}^{+\lambda_m} \frac{1}{L^2} \cdot \frac{\partial}{\partial u} \left\{ \frac{(l+h)^2 \cdot w_{\theta_n}(\lambda, u)}{\sqrt{(l+h)^2 + (\lambda-u)^2}} \cdot p_{\theta_n}(\lambda, u) \right\} \bigg|_{u=u^*} d\lambda, \quad (3)$$

$$Sb_\eta^{\theta_n}(\vec{x}) = \frac{1}{2} \int_{-u_m}^{+u_m} \frac{1}{H^2} \cdot \frac{\partial}{\partial \lambda} \left\{ \frac{(l+h)^2 \cdot w_{\theta_n}(\lambda, u)}{\sqrt{(l+h)^2 + (\lambda-u)^2}} \cdot p_{\theta_n}(\lambda, u) \right\} \bigg|_{\lambda=\lambda^*} du, \quad (4)$$

where $\vec{x} = (x, y)$ is the reconstructed point; $L = -x\sin\theta_n + y\cos\theta_n + l$, $H = x\sin\theta_n - y\cos\theta_n + h$; $u^*$ is the local coordinate of the ray from the source focus $\lambda$ to passing through $\vec{x}$ on the detector, $\lambda^*$ is the local coordinate of the ray from the detector element $u$ to passing through $\vec{x}$ on the source trajectory. $p_{\theta_n}(\lambda, u)$ is the projection of the *n*-th STCT. $w_{\theta_n}(\lambda, u)$ is a weighting function to eliminate the redundancy data in mSTCT, whose details can be found in the paper [19]. To transform from the DBP image to the object space $f_{\theta_n}(\vec{x})$, the inversion of Hilbert transform in the second step needs to be performed for each DBP image along the direction parallel to the source translation [20]. $\eta$ denotes the angle between the filtering lines of Hilbert transform and the positive *y*-axis. Finally, overlapping all $f_{\theta_n}(\vec{x})$ to complete reconstruction within the FOV, i.e., $f_\theta = \sum_{n=1}^{T} f_{\theta_n}(\vec{x})$.

The previous experiments demonstrated that D-BPF achieved high-quality reconstruction under the same fewer projections compared to S-BPF and V-FBP. Since the differential is equivalent to filtering along the detector in D-BPF, its cutoff frequency is influenced by the size and number of elements, whereas others are along the source trajectory and are influenced by the source sampling points.

### 3. Method

#### *3.1 Problem analysis and strategy*

In the mSTCT reconstruction algorithms, our previously proposed D-BPF is verified to be more practical as it significantly improves scanning efficiency and avoids truncation. However, we notice that the reconstructed results within the enlarged FOVs show artifacts. As shown in Figure 2, these artifacts are exacerbated as the FOV extends, which deviates from the intention of mSTCT imaging. Not only do unstable values appear at the edges, but other artifacts appear in the reconstructed image as the FOV expands. It is significant to improve the reconstruction accuracy under the extreme enlarged FOV, and it may lead to important progress in this field.

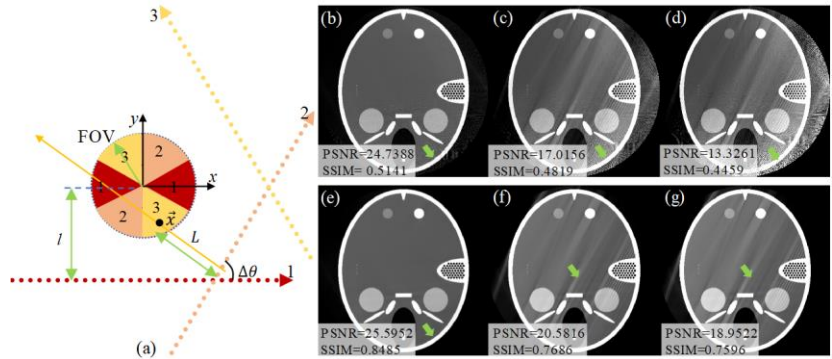

Fig. 2. Reconstruction results with the different enlarged FOVs via D-BPF and S-BPF for the half-scan mSTCT imaging geometry consisting of the three-segment STCTs: (a) mSTCT imaging model consists of three-segment STCTs, and the interval between adjacent STCTs is about 62.8°; (b)–(d) reconstructed images when $2\lambda_m$ is set to 40 mm, 45 mm, and 50 mm via D-BPF with the number *N* of source sampling points per STCT being 251, the corresponding

radius of the enlarged FOV is 8.3921 mm, 10.1 mm, and 11.77 mm (the half-sizes of the objects are set to them), respectively; (b)–(d) reconstructed images via S-BPF under the same scanned conditions as (b)–(d) with $N$ being 1001. The reason for the selection of different source sampling points (251 and 1001, respectively) for D-BPF and S-BPF can be found in our previous research [20]. A display window is [0, 3].

**Analysis:** In F-mSTCT, when not considering redundancy between adjacent STCTs, each reconstructed point is equivalent to being measured twice by complementary rays. Therefore, the errors caused by the factor $1/L^2$ related to the distance from the reconstructed point to the source may be balanced by a value of one far and one near [25,32]. In the extreme enlarged FOV, there are small values close to zero in the region closer to the source, which makes $1/L^2$ show unstable, intolerable errors and may not be well balanced. Unlike D-DBP, the factor in S-DBP is $1/H^2$, where $H$ denotes the vertical distance from the reconstructed point $\vec{x}$ to the detector. Therefore, the errors generated in S-BPF are theoretically smaller and may be suitable for F-mSTCT reconstruction under the large FOV. Besides, the redundancy weighting established cannot adjust as the FOV of mSTCT extends, so artifacts appear within the reconstructed image from Figure 2.

**Strategy:** Attempt to explore the effectiveness and applicability of D-BPF and S-BPF in F-mSTCT mode (see Figure 1(c)). The redundancy of F-mSTCT is different from mSTCT, in F-mSTCT, each point to be reconstructed is covered by 360°. However, due to the intersection between source trajectories and detectors, the projections of adjacent STCTs overlap, resulting in redundant projection data and affecting the final reconstruction results. Therefore, we need to derive a redundancy weighting method that is suitable for F-mSTCT and can be adaptively adjusted as the FOV extends. Subsequently, we apply it to D-BPF and S-BPF.

*3.2 Redundancy-weighted strategy for F-mSTCT*

Inspired by the redundancy-weighted idea in symmetric-geometry computed tomography [34], we analyze and derive the redundant situation and obtained a new redundancy-weighted function applicable to F-mSTCT.

According to the geometric relationship shown in Figure 3, the projection data $p_{\theta_n}(\lambda, u)$ of the $n$-th STCT can be mapped to the equivalent virtual detector that is parallel to the detector past the center of rotation and denoted as $q_{\theta_n}(\lambda, t)$ one by one. The mapping relationship between $p_{\theta_n}(u, \lambda)$ and $q_{\theta_n}(t, \lambda)$ is shown in Eq. (5):

$$p_{\theta_n}(\lambda, u) = q_{\theta_n}(\lambda, t), \qquad u = \frac{l+h}{l}t - \frac{h}{l}\lambda. \tag{5}$$

Here, $t$ is the one-dimensional row coordinate of the x-ray on the equivalent virtual detector.

The redundant projection per STCT in F-mSTCT is the same, and the schematic of the redundant projection data of adjacent STCTs is shown in Figure 3.

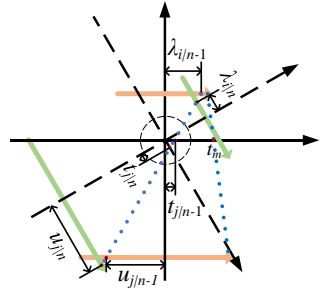

Fig. 3. Geometric schematic of the redundant projection of two adjacent STCTs.

Let the range of sampling points of the ray source trajectory be $\pm\lambda_m$, the range of detector row coordinates be $\pm u_m$, and the maximum range of the virtual detector be $\pm t_m$. $t_m$ is the intersection of the ray represented by $(\lambda_m, u_m)$ with the $t$-axis, the maximum value to which $t$ goes, and $-t_m$ is the same as this. All projections of each STCT segment are located within the rectangular coordinate range of $[-t_m, t_m]$ and $[-\lambda_m, \lambda_m]$, and the projection data will fill the whole rectangular coordinate range formed by $[-t_m, t_m]$ and $[-\lambda_m, \lambda_m]$ when the detector length limitation is not considered.

Then, we mapped the projection data in the $t - \lambda$ coordinate space to Radon space and redistributed the angular intervals of each STCT segment in Radon space according to the full-scan requirement to obtain the redundancy overlap region between the projection data of two adjacent STCT segments. As shown in Figure 4(a), in the $t - \lambda$ coordinate space, the regions $A$ and $B$ of the projection data of the previous STCT segment overlap with the regions $A'$ and $B'$ of the later STCT segment, respectively. In addition, a more detailed description of Figure 4 is in the **Supplement**.

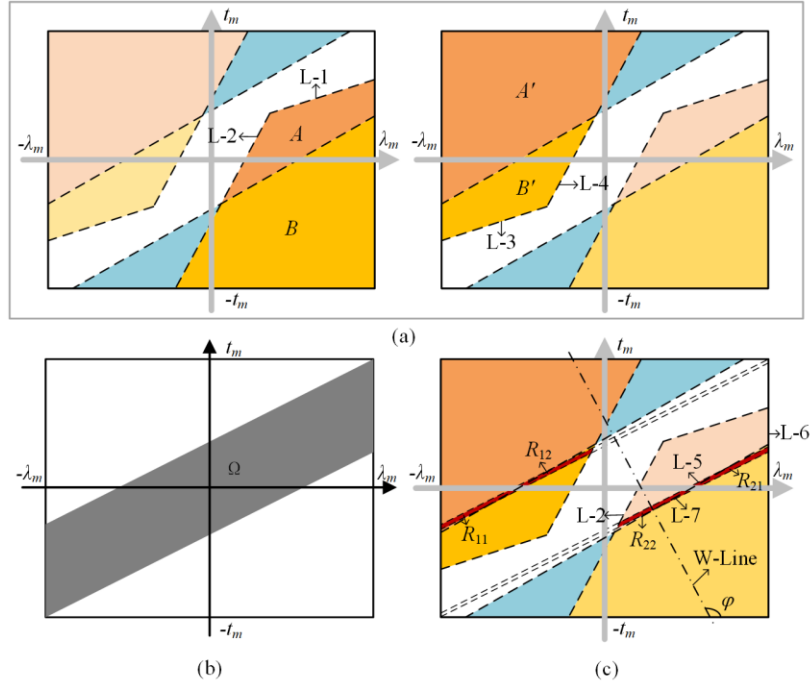

Fig. 4. Analysis of the redundant situation for F-mSTCT: (a) a redundant case of two adjacent STCTs in $t - \lambda$ coordinate space, (b) the effective projection region $\Omega$ for each STCT in the $t - \lambda$ coordinate space, and (c) redundancy-weighted analysis of projection data for STCT.

For the mutually overlapping pair of projection data $q_{\theta_{n-1}}(\lambda_{i|n-1}, t_{j|n-1})$ and $q_n(\lambda_{i|n}, t_{j|n})$ in the redundant regions $A$, $B$, $A'$, $B'$ shown in Figure 4(a), there exists the relationship as shown in Eq. (6):

$$q_{\theta_{n-1}}(\lambda_{i|n-1}, t_{j|n-1}) = q_{\theta_n}(\lambda_{i|n}, t_{j|n}). \tag{6}$$

Here, the algebraic relationship between the overlapping projection coordinates $(\lambda_{i|n-1}, t_{j|n-1})$ and $(\lambda_{i|n}, t_{j|n})$ shown in Eq. (6) satisfies Eq. (7). (The detailed derivation of Eq. (7) is shown in the **Supplement**).

$$\begin{cases} t_{i|n-1} = t_{j|n}\sqrt{\dfrac{\left[\dfrac{(\lambda_{i|n}-t_{j|n})l+l^2\tan\dfrac{\theta}{2}}{l-(\lambda_{i|n}-t_{j|n})\tan\dfrac{\theta}{2}}\right]^2+l^2}{(\lambda_{i|n}-t_{j|n})^2+l^2}} \\ \lambda_{i|n-1} = t_{j|n-1} + \dfrac{(\lambda_{i|n}-t_{j|n})l+l^2\tan\dfrac{\theta}{2}}{l-(\lambda_{i|n}-t_{j|n})\tan\dfrac{\theta}{2}} \end{cases}. \qquad (7)$$

Due to the finite length detector, the projection within the rectangular coordinate range formed by $[-t_m, t_m]$ and $[-\lambda_m, \lambda_m]$ in each STCT do not all exist. Only the projection data located within the rectangular coordinate range of $[-u_m, u_m]$ and $[-\lambda_m, \lambda_m]$ in the $u-\lambda$ coordinate space is valid and is mapped to the area $\Omega$ in the $t-\lambda$ coordinate space. Area $\Omega$ is called the valid projection (see Figure 4(b)).

Combining with the valid projection region (Figure 4(b)), the redundant region of the projection data of one STCT can be obtained, as shown in Figure 4(c), where the region $R_1$ (including $R_{11}$ and $R_{12}$) is the overlapped redundant region of this segment STCT with the previous segment STCT, and the region $R_2$ (including $R_{21}$ and $R_{22}$) is the overlapped redundant region of this segment STCT with the next segment STCT. For the mutually overlapping pair of projection data $q_{\theta_{n-1}}(\lambda_{i|n-1}, t_{j|n-1})$ and $q_n(\lambda_{i|n}, t_{j|n})$ in the redundant regions $R_{11}, R_{12}, R_{21}, R_{22}$ shown in Figure 4(c), they also satisfy the relationship as shown in Eq. (6) and Eq. (7) as well.

Once the redundant region is determined, the corresponding redundancy-weighted function can be designed. According to the literature [22], the redundancy-weighted function needs to satisfy two conditions: 1) In the redundant region, each point to be reconstructed has only one equivalent x-ray passing through it; 2) The weighting function must be continuous.

As shown in Figure 4(c), the W-Line is the directional line of redundancy weighting, whose slope $\tan\varphi$ determines the direction of redundancy weighting, and L-2, L-5, L-6, and L-7 are the boundary lines between the redundant region $R_2$ and the surrounding non-redundant regions. The redundancy-weighted function of the $R_2$ region is determined by the distance of the projection data points along the direction of the W-Line to the four boundary lines L-2, L-5, L-6, and L-7, and the redundancy-weighted function value of the redundant region $R_1$ and its corresponding redundancy-weighted function value of the projection data points in the $R_2$ region in the previous STCT sum to one. In summary, the redundancy-weighted function for the *n*-th STCT projection data is shown in Eq. (8).

$$w_{\theta_n}(\lambda_{i|n}, t_{j|n}) = \begin{cases} 1, & (\lambda_{i|n}, t_{j|n}) \notin [R_1, R_2] \\ f\left(\dfrac{\min(dis_{w-2}, dis_{w-5})}{\min(dis_{w-2}, dis_{w-5})+\min(dis_{w-6}, dis_{w-7})}\right), & (\lambda_{i|n}, t_{j|n}) \in [R_2] \\ 1 - w_{\theta_{n-1}}(\lambda_{i|n-1}, t_{j|n-1}), & (\lambda_{i|n}, t_{j|n}) \in [R_1] \end{cases}.$$

(8)

Here, $dis_{w-2}$ denotes the distance of the current projection data point $(\lambda_{i|n}, t_{j|n})$ along the direction of the W-Line to the L-2 boundary line, and $dis_{w-5}$, $dis_{w-6}$, and $dis_{w-7}$ are the same way. That is, the length of the W-Line passing through the current projection data point $(\lambda_{i|n}, t_{j|n})$ in the redundant region $R_2$ is obtained as the denominator, and the distance of the point $(\lambda_{i|n}, t_{j|n})$ along the direction of the W-Line to the dividing lines L-2 and L-5 is calculated, the smallest of which is taken as the numerator. The ratio obtained above is smoothed by the function $f(x)$ shown in Eq. (9) to finally obtain the redundancy weighting function for the redundant region $R_2$. The redundancy-weighted function of the $R_1$ region is obtained by calculating the redundancy-weighted function of the $R_2$ region of the previous section of STCT corresponding to it. $(\lambda_{i|n-1}, t_{j|n-1})$ and $(\lambda_{i|n}, t_{j|n})$ in Eq. (8) satisfy Eq. (7).

$$f(x) = \begin{cases} 1, & x \leq 0 \\ \dfrac{1 + \sin((0.5 - x)\pi)}{2}, & 0 < x < 1 \\ 0, & x \geq 1 \end{cases} \quad (9)$$

To illustrate the generation procedure of the proposed redundancy-weighted function and replace $w_{\theta_n}(\lambda, u)$ in the reconstruction formulas (Eqs. (3) and (4)) above more visually, Table 1 lists the pseudo-code. Figures 5(a) and (b) show the weighting results of one STCT projection in the $t$-$\lambda$ coordinate space and the corresponding $u$-$\lambda$ coordinate space, respectively.

**Table 1. The pseudo-code for calculating the redundancy-weighted function**

Redundancy-weighted function of the *n*-th STCT

1: **Input:** $\hat{w}_\theta = 1$, $w_\theta = 1$, $s, d$.
2: **For** $\lambda_{i|n} = -\lambda_m : \lambda_m$ do
3:    **For** $u_{j|n} = -u_m, u_m$ do
4:       Calculate $t_{j|n}$ by Eq. (5);
5:       **If** $(\lambda_{i|n}, t_{j|n}) \in [R_2]$
6:          Calculate $dis_{w-2}\ dis_{w-5}\ dis_{w-6}\ dis_{w-7}$;
7:          Calculate $w_{\theta_n}(\lambda_{i|n}, t_{j|n})$ by Eq. (8);
8:       **Elseif** $(\lambda_{i|n}, t_{j|n}) \in [R_1]$
9:          Calculate $(\lambda_{i|n-1}, t_{j|n-1})$ by Eq. (7);
10:         Calculate $dis_{w-2}\ dis_{w-5}\ dis_{w-6}\ dis_{w-7}$;
11:         Calculate $w_{\theta_n}(\lambda_{i|n-1}, t_{j|n-1})$ by Eq. (8);
12:         Calculate $w_{\theta_n}(\lambda_{i|n}, t_{j|n})$ by Eq. (8);
13:       $w_{\theta_n}(\lambda_{i|n}, u_{j|n}) = w_{\theta_n}(\lambda_{i|n}, t_{j|n})$;
14:    **End for**
15: **End for**
16: **Output:** $w_{\theta_n}(\lambda_{i|n}, u_{j|n})$.

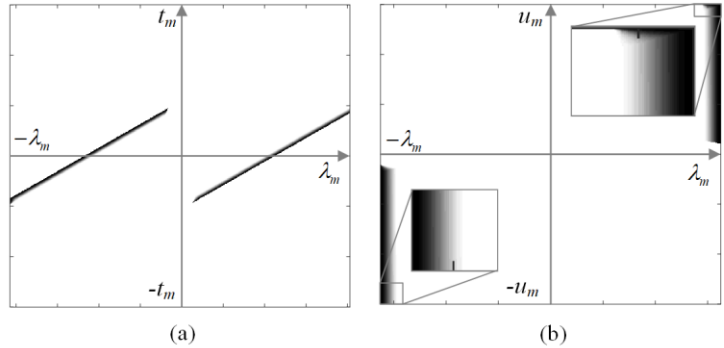

Fig. 5. The weighting results when the weighting direction is $\tan\varphi = 3\pi/4$: (a) the weighting results of one STCT projection data in the $t - \lambda$ coordinate space; (b) the weighting results of one STCT projection data in the corresponding $u - \lambda$ coordinate space. A display window is [0, 1].

## 4. Experiments

Some numerical experiments are conducted to evaluate our methods. These methods are implemented based on the Astra-Python software tool, and conducted on a computer with an Intel (R) Core i7-8550U CPU @ 1.80 GHz and an NVIDIA GeForce MX250. Table 2 lists scanning parameters, where the length of source translation $2\lambda_m$ is set to 40 mm to obtain an enlarged FOV with a radius $R$ of 8.3921 mm (Eq. (2)), which exceeds the FOV with a radius $r$ of 6.5402 mm in the standard CT. Additionally, the number $T$ of STCTs in F-mSTCT can be determined to be six, and it is relatively three in mSTCT. The FORBILD phantom, with the size being $2R \times 2R$ and the reconstructed size being $512 \times 512$ pixels, is used to test.

Table 2. Scanning parameters

| Parameters | Value |
| --- | --- |
| Source-to-origin distance $l$ (mm) | 13.75 |
| Detector-to-origin distance $h$ (mm) | 106.5 |
| Length of source translation $2\lambda_m$ (mm) | 40 |
| Source sample points per STCT $N$ | 251 (D-BPF, FD-BPF, and FW-FBP); 1001 (S-BPF, FS-BPF) |
| Number of detector elements $J$ | 1024 |
| Size of detector element (mm$^2$) | 0.127×0.127 |
| Number of STCTs in F-mSTCT $T$ | 6 |
| Interval translation angle $\Delta\theta$ (°) | 60 |
| Radius of FOV $R$ (mm) | 8.3921 |

As shown in Figure 6, this experiment can verify effectiveness that the proposed redundancy-weighted function for the F-mSTCT reconstruction.

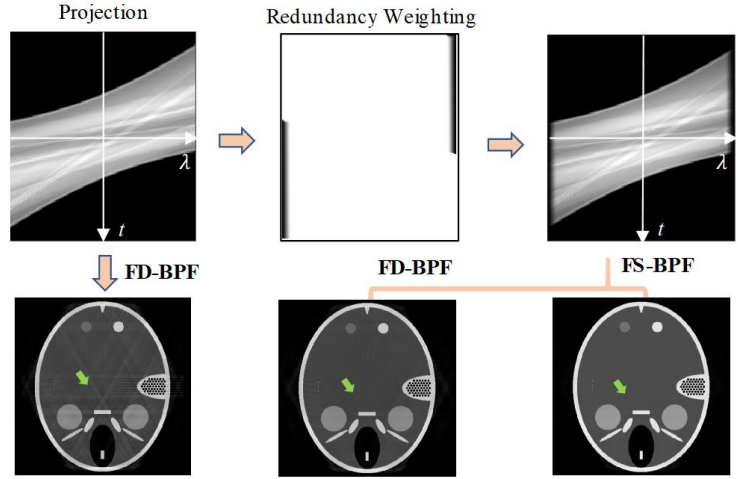

Fig. 6. Effectiveness illustration of the redundancy-weighted function for the projection in F-mSTCT reconstruction. The display window is [0, 3].

To investigate the performance of FD-BPF and FS-BPF, some experiments are performed under different FOVs, as illustrated in Figures 7(a)–(c). More specifically, we set $2\lambda_m$ to 40 mm, 45 mm, and 50 mm, i.e., the radius of the enlarged FOV is 8.39 mm, 10.1 mm, and 11.77 mm, and the magnification of the FOV is 1.28, 1.54, and 1.8 compared to the standard circular scanning, respectively. In F-mSTCT, we can equate it to detector bias and combine it with smooth redundancy weights to weaken truncation. For this reason, we introduce full scan-weighted FBP (abbreviated as FW-FBP) as a comparison. In the three different FOVs, the reconstructed results, including images and profiles via FW-FBP, FD-BPF, and FS-BPF, are shown in Figures 9(d)–(g), (h)–(k), and (l)–(o), respectively. The two kinds of metrics, including the power signal-to-noise ratio (PSNR) and structural similarity (SSIM), are located at the left-bottom corner of each reconstructed image.

With the radius of the enlarged FOV being 8.39 mm, FW-FBP can achieve high-quality reconstruction, as shown in Figures 9(d) and (g). However, as the FOV continuously increases, the images reconstructed through FW-FBP exhibit significant intensity dropouts within the FOV (Figure 7(g)). Moreover, taking Figures 2(e)–(g), which show the reconstructed images of D-BPF as references, the reconstruction result of FD-BPF, as originally envisioned, can alleviate the problem of large reconstruction errors. However, as the size of the tested object and FOV increase, some sharp streak artifacts still appear in the reconstructed area near the FOV boundary (Figures 9(i) and (j)), which are numerically unstable and sharp (Figure 7(k)) and affect the imaging quality. Thus, FD-BPF still cannot meet the urgent need for high-quality imaging of large objects under large FOVs. In contrast, as shown in Figures 7(l)–(o), FS-BPF can reconstruct high-quality images without visible artifacts, and the profile from 0 to 100 pixels in the central row is close to the truth, even though it requires more projection data ($N = 1001$) than the two algorithms ($N = 251$) mentioned above. In high-quality imaging, the issue of reduced efficiency caused by acquiring a large amount of projection data can be accepted.

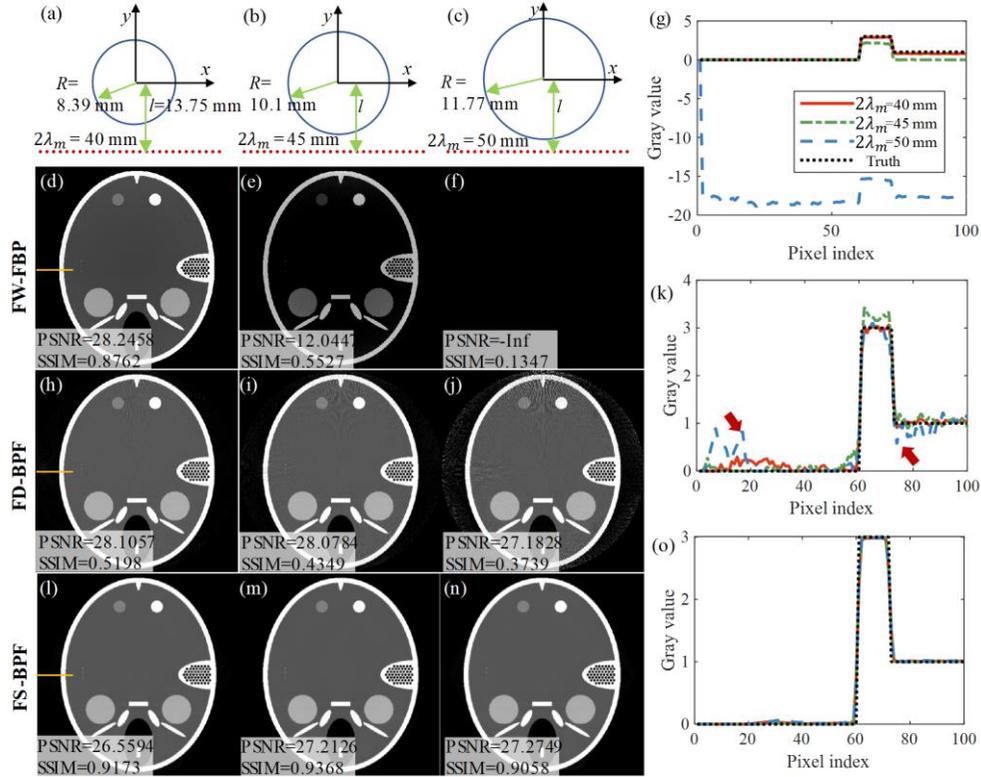

Fig. 7. Reconstructed results via FW-FBP, FD-BPF, and FS-BPF under the different enlarged FOVs: (a)-(c) imaging diagrams of different enlarged FOVs; (d)-(f) reconstructed images via FW-FBP under the conditions of (a)-(c); (g) profiles from 0 to 100 pixels in the horizontal center of (d)-(f); (h)-(j) reconstructed images via FD-BPF; (g) profiles of (h)-(j); (l)-(n) reconstructed images via FS-FBP; and (o) the profiles of (h)-(j). The display window is [0, 3].

To further investigate the performance of FS-BPF, we continuously enlarge the FOV and the measured object to let the reconstructed region be extremely close to the source trajectory, as described in Figures 8(a)–(c). More specifically, $2\lambda_m$ is set to 52 mm, 54 mm, and 55 mm, corresponding to the radius of the FOV being 12.4318 mm, 13.0855 mm, and 13.4102 mm, and the magnification of the FOV is about 1.9, 2, and 2.05 compared to the standard circular scanning, respectively. Figures 8(d)–(f) display the reconstructed images and PSNR value corresponding to the conditions of Figures 8(a)–(c), and Figure 8(g) shows the profile from 350 to 512 pixels in the horizontal center of three images. Besides, we also choose a medical CT slice with 512 × 512 pixels as the tested object. With $N$ being 1201, the high-quality reconstructed results are presented in Figures 8(h)–(k) without visible artifacts, even though the reconstructed regions are extremely close to the source.

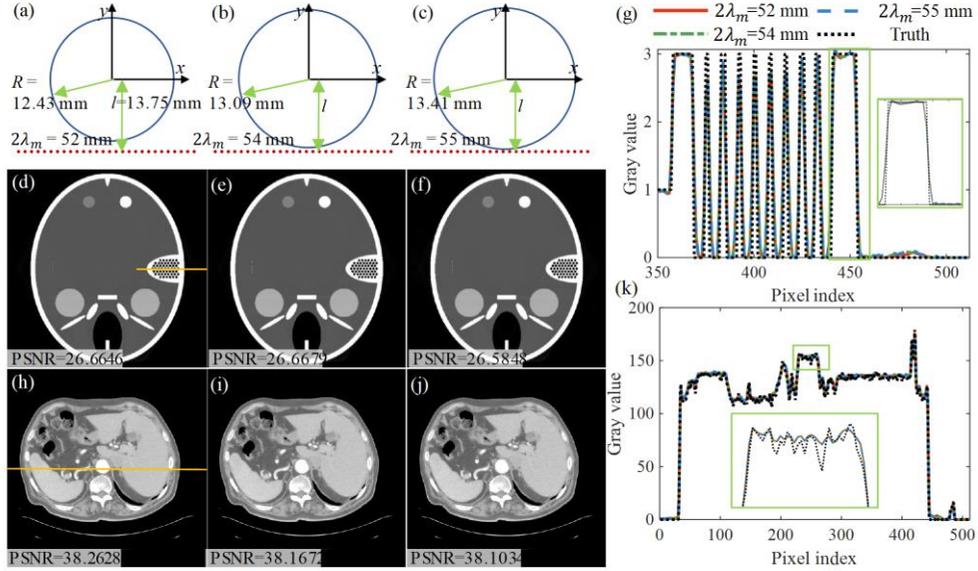

Fig. 8. Reconstructed results via FS-BPF under the extended FOVs where the reconstruction area is very close to the source trajectory: (a)-(c) imaging diagrams of different enlarged FOVs; (d)-(f) reconstructed images under the conditions of (a)-(c); (g) profiles from 350 to 512 pixels in the horizontal center of (d)-(f); (h)-(j) reconstructed images for a medical phantom; and (k) horizontal central profiles of (h). The window ranges are [0, 3] and [90, 150] (HU) for the FORBILD phantom and medical phantom, respectively.

## 5. Discussion

Aiming at solving the problem of reconstruction errors under the large FOVs of mSTCT and extending the FOV as much as possible to achieve high-quality imaging, we propose FD-BPF and FS-BPF by constructing F-mSTCT and corresponding to the redundancy weights.

When the FOV is extended infinitely close to the source, FS-BPF still obtain satisfactory reconstructed images even though it requires a large number of projections. Impressively, when the FOV is extended to 2.05 times, the reconstruction results of FS-BPF remain stable. In fact, the result can demonstrate the F-mSTCT reconstruction breaks through the theoretical maximum limit of two times for detector bias expansion FOV imaging methods (the two times are only theoretical values, and practical high-quality imaging is less than two times due to the requirement for enough redundancy).

From the numerical experimental results, we can preliminarily conclude the following:

1) When the amplification of the FOV only needs to be less than about 1.28 (i.e., the FOV is slightly enlarged), we can use FW-FBP to reconstruct and keep scanning efficiency;

2) When amplifying FOV requires a range of 1.28–1.54 times and higher scanning efficiency is needed, FD-FBP can be used to reconstruct;

3) When amplifying FOV exceeds 1.54 times and there is no concern for scanning efficiency, FS-BPF can be used to reconstruct better-quality images.

*5.1 Differences on the redundancy weighting for mSTCT and F-mSTCT*

Compared with the redundancy-weighted function for mSTCT, our proposed redundancy weighting method has the following advantages:

1) It has a simpler redundancy profile with only one type of redundancy, whereas the proposed method in mSTCT [19] has three types of redundancy and a more complex redundancy profile;

2) The redundancy from the geometric relationship between the ray source and the detector was analyzed in mSTCT, while the redundancy of the projection after mapping the projection to the Radon space was analyzed in our proposed method, which makes the process of deriving the redundant region of projection data more accurate and clear;

3) We optimize the weighting of the redundant projection data in such a way that the direction of the redundancy weighting can be adjusted, which cannot be achieved by the previous weighting method [19] (see reconstructed results in Figure 2). More specifically, the paper [19] determined the projection redundancy in $u$-$\lambda$ coordinate space based on the location of the intersection of one STCT projection line with the source trajectory and detector of an adjacent STCT. It is proven that the redundant region extraction method described in this paper is basically consistent with the redundant regions extracted by this method, and both achieve quite accurate redundant region extraction. Additionally, when designing the redundancy-weighted function, its method is to calculate the redundancy-weighted function by directly calculating the distance from the projection coordinate point to the corresponding redundancy boundary, and the weighting direction and effect are fixed, while in this paper, a W-Line is added as the weighting direction line, which makes the weighting direction adjustable and thus ensures the smooth continuity of the redundancy-weighted function at the boundary under different scanning parameters and then obtains a better reconstruction effect.

*5.2 Analysis of reconstruction differences among these methods*

Taking F-mSTCT to be equivalent to the detector bias system, FW-FBP utilizes the idea of the weighted FBP to weaken the truncation. Due to a global ramp filter, the truncation error will affect the whole filtered data if the smooth curve for redundant projection is narrow. When $2\lambda_m \leqslant 40$ mm, FW-FBP gets competitive results, whereas its errors significantly increase with $2\lambda_m$ being greater than 40 mm. The phenomenon of two boundaries being simultaneously truncated may occur with the enlarged size of the FOV and object. In this case, directly using the weighted FBP idea of detector bias (usually truncation only occurs on one side) may not be suitable.

FD-BPF can weaken the unstable errors in areas where the reconstructed points are close to the source. However, as the FOV and reconstructed object are extremely enlarged, the distance parameter $L$ in the backprojection weighting factor $1/L^2$ approaches zero in the region near the source, which leads to highly unstable weighted values that are difficult to be balanced by complementary rays.

In the reconstructed area near the source, the distance parameter $H$ in the backprojection weighting factor $1/H^2$ of FS-BPF is much larger than $L$ due to the high magnification. Therefore, the weighting factor $1/H^2$ has a weak impact on reconstruction errors. Additionally, the projection of one STCT is non-truncated along the source trajectory, which will not produce errors at the truncation points after the finite difference like FD-BPF. Even with the requirement of large projection data, FS-BPF can obtain high-quality images when the FOV is extremely extended and close to the source.

## 6. Conclusion

In the half-scan mSTCT reconstruction, the backprojection weighting factors cause artifacts and errors as the FOV and object size increase. Besides, the existing redundancy weighting for mSTCT is not working well with adjusting FOVs. To this end, we combine the F-mSTCT geometry with the BPF algorithms to propose FD-BPF and FS-BPF, and design a redundancy-weighted function that can adaptively adjust according to the F-mSTCT geometry. The experiments demonstrate that FS-BPF has the ability to reconstruct high-quality images using large amounts of projection data, even if the extended FOV is extremely close to the source. Finally, according to the different practical requirements for extended FOVs, we provide suggestions on how to choose a reasonable analytical algorithm.

**Acknowledgments.** This research work is supported by Nationa National Natural Science Foundation of China (52075133) and CGN-HIT Advanced Nuclear and New Energy Research Institute (CGN-HIT202215). Authors thank Prof. Fenglin Liu and Dr. Haijun Yu in the Engineering Research Center of Industrial Computed Tomography Nondestructive Testing, Ministry of Education, Chongqing University, for selflessly providing me with learning materials on mSTCT and its reconstruction algorithms.

**Disclosures.** The authors declare no conflicts of interest.

**Data availability.** Data underlying the results presented in this paper can be obtained from the authors upon reasonable request.

### *Supplementary*: *Details on calculating redundancy weights*

In this subsection, the derivation process of the overlapping redundant region between two adjacent STCTs shown in Figure 4 is analyzed in detail, and the relationship shown in Eq. (7) is explained as well.

Equating the fan-beam projection data $q_{\theta_n}(\lambda, t)$ of each STCT to the parallel-beam projection data $q'_{\theta_n}(\alpha, s)$, where $\alpha$ is the angle of the rebinning-to-parallel-beam detector and $s$ is the coordinate of it. The relationship between $(\lambda, t)$ and $(\alpha, s)$ is as follows:

$$\begin{cases} \alpha = \arctan\dfrac{\lambda - t}{l} - \theta_n \\ s = \dfrac{lt}{\sqrt{(\lambda - t)^2 + l^2}} \end{cases}, \quad (S1)$$

where $\theta_n$ denotes the rotation angle of the *n*-th STCT.

The fan-beam projection data $q_{\theta_n}(\lambda, t)$ of each STCT can be mapped into the Radon space using the relationship shown in Eq. (S1), as shown in Figure S1.

To ensure that each point to be reconstructed is covered by 360° and to reduce redundancy, the projection angle range assumed by each STCT in Radon space is $2\pi/T$, where $T$ is the total number of STCTs in F-mSTCT. The specific projection angle zone $\Delta_\alpha$ assumed by each STCT can be obtained by combining the specific rotation direction with the geometric relationship. The specific projection angle zone $\Delta_\alpha$ assumed by each STCT can be obtained by combining the specific rotation direction with the geometric relationship. Figure S1 shows the distribution of the projection angle zone when the total number of STCTs is six and the x-ray source is rotated equivalently clockwise.

As shown in Figure S1, the projection regions $R_1^n$ and $R_2^n$ beyond the angular range assumed by this STCT are considered the redundant regions. Combining the relationship between $(\lambda, t)$ and $(\alpha, s)$ shown in Eq. (S1), the $R_1^n$ region and the $R_2^n$ region are formulated as follows:

$$R_1^n: (\lambda_{i|n}, t_{j|n}) \begin{cases} \lambda_{i|n} - t_{j|n} \geq l \tan\dfrac{\pi}{T}, \\ (\lambda_{i|n}, t_{j|n}) \in \Omega \end{cases} \quad (S2)$$

$$R_2^n: (\lambda_{i|n}, t_{j|n}) \begin{cases} \lambda_{i|n} - t_{j|n} \leq l \tan\left(-\dfrac{\pi}{T}\right) \\ (\lambda_{i|n}, t_{j|n}) \in \Omega \end{cases}. \quad (S3)$$

The final redundant region $R_{11}$ is obtained by adding the effective projection range limits of the adjacent STCTs that overlap with this STCT to $R_1^n$. The relationship between the $R_2^n$ region and the $R_{21}$ region is similar to this. The $R_{11}$ and $R_{21}$ regions can be expressed as follows:

$$R_{11}: (\lambda_{i|n}, t_{j|n}) \begin{cases} \lambda_{i|n} - t_{j|n} \geq l \tan\dfrac{\pi}{T} \\ (\lambda_{i|n}, t_{j|n}) \in \Omega \\ (\lambda_{i|n-1}, t_{j|n-1}) \in \Omega \end{cases}, \quad (S4)$$

$$R_{21}: (\lambda_{i|n}, t_{j|n}) \begin{cases} \lambda_{i|n} - t_{j|n} \leq l \tan\left(-\dfrac{\pi}{T}\right) \\ (\lambda_{i|n}, t_{j|n}) \in \Omega \\ (\lambda_{i|n+1}, t_{j|n+1}) \in \Omega \end{cases}, \quad (S5)$$

where $(\lambda_{i|n-1}, t_{j|n-1})$ and $(\lambda_{i|n}, t_{j|n})$ and $(\lambda_{i|n+1}, t_{j|n+1})$ in Eqs. (S4) and (S5) satisfy Eq. (7).

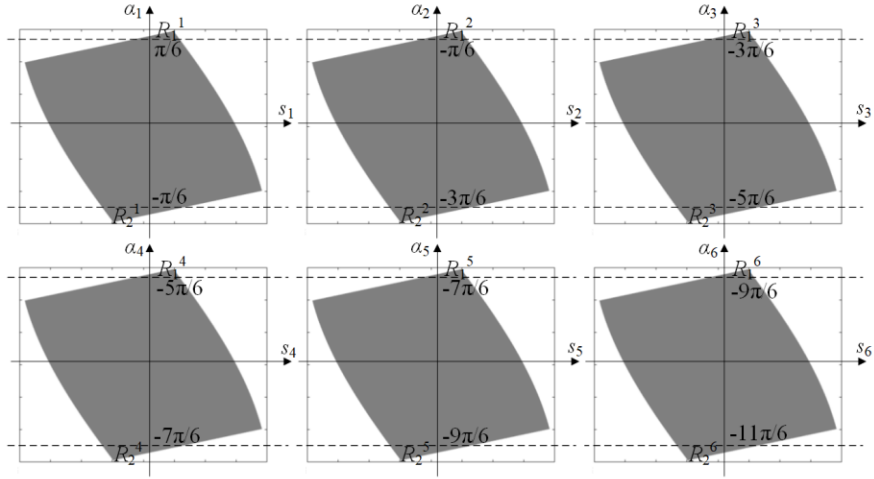

Fig. S1. Distribution of projection angle zones for a set of specified parameters. The total number of STCTs is six, and the x-ray source is rotated equivalently clockwise.

Combining with Eq. (7), it is obtained that the $R_{11}$ region overlaps with the $R_{22}$ region of the previous STCT, and the $R_{21}$ region overlaps with the $R_{12}$ region of the next STCT. In summary, the projection redundant region of each STCT can be obtained as shown in $R_1$ and $R_2$ in Figure 4.

Next, we elaborate on the algebraic relationship between the projection data within the redundant regions of the two adjacent STCTs shown in Eq. (7).

In F-mSTCT, there is only one case of redundancy in the projection data, i.e., the case where the x-rays of two adjacent STCTs overlap, as shown in Figure 3, where the following relationship exists between the projections $q'_{\theta_n}(\alpha_{\langle i,j \rangle|n}, s_{\langle i,j \rangle|n})$ and $q'_{\theta_{n+1}}(\alpha_{\langle i,j \rangle|n+1}, s_{\langle i,j \rangle|n+1})$ of the two adjacent STCTs:

$$\begin{cases} \alpha_{\langle i,j \rangle|n} = \alpha_{\langle i,j \rangle|n+1} \\ s_{\langle i,j \rangle|n} = s_{\langle i,j \rangle|n+1} \end{cases}. \quad (S6)$$

Combining Eq. (S1), we can obtain from Eq. (S6):

$$\begin{cases} \arctan\dfrac{\lambda_{i|n} - t_{j|n}}{l} - \theta_n = \arctan\dfrac{\lambda_{i|n+1} - t_{j|n+1}}{l} - \theta_{n+1} \\ \dfrac{l \cdot t_{j|n}}{\sqrt{(\lambda_{i|n} - t_{j|n})^2 + l^2}} = \dfrac{l \cdot t_{j|n+1}}{\sqrt{(\lambda_{i|n+1} - t_{j|n+1})^2 + l^2}} \end{cases}, \quad (S7)$$

then Eq. (7) can be obtained from Eq. (S7).